\documentclass[10pt, a4paper]{article}
\usepackage{lrec2016}
\usepackage{url}
\usepackage{booktabs}
\usepackage{multicol}
\usepackage{graphicx}
\usepackage[latin1]{inputenc}
\usepackage{enumitem}
\setlist[enumerate]{itemsep=0px, parsep=0px}
\setlist[description]{itemsep=2px, parsep=0px, leftmargin=0cm, topsep=2px}

\title{Stereotyping and Bias in the Flickr30K Dataset}
\name{Emiel van Miltenburg}
\address{Vrije Universiteit Amsterdam\\emiel.van.miltenburg@vu.nl}
\usepackage[usenames,dvipsnames]{xcolor}
\abstract{\raisebox{4.5cm}[0px]{\makebox[0px][l]{\parbox{\textwidth}{\footnotesize\centering\color{BrickRed} In: Proceedings of the Workshop on Multimodal 
Corpora: Computer vision and language processing (MMC-2016), pages 1--4.\\Workshop held: 24 May 2016, collocated with LREC 2016, Portoro\v z, Slovenia.\\Proceedings available at: \url{http://www.lrec-conf.org/proceedings/lrec2016/workshops/LREC2016Workshop-MCC-2016-proceedings.pdf}}}}An untested assumption behind the crowdsourced descriptions of the images in the Flickr30K dataset \cite{young2014image} is that they ``focus only on the information that can be obtained from the image alone'' \cite[p. 859]{hodosh2013framing}. This paper presents some evidence against this assumption, and provides a list of biases and unwarranted inferences that can be found in the Flickr30K dataset. Finally, it considers methods to find examples of these, and discusses how we should deal with stereotype-driven descriptions in future applications.\\
\newline
}

\begin{document}

\maketitleabstract

\section{Introduction}
The Flickr30K dataset \cite{young2014image} is a collection of over 30,000 images with 5 crowdsourced descriptions each. It is commonly used to train and evaluate neural network models that generate image descriptions (e.g.\ \cite{vinyals2015show}). An untested assumption behind the dataset is that the descriptions are based on the images, and nothing else. Here are the authors (about the Flickr8K dataset, a subset of Flickr30K):

\begin{quote}
``By asking people to describe the people, objects, scenes and activities that are shown in a picture without giving them any further information about the context in which the picture was taken, we were able to obtain conceptual descriptions that focus only on the information that can be obtained from the image alone.''
\cite[p. 859]{hodosh2013framing}
\end{quote}

\begin{figure}[h!]
\centering
\includegraphics[width=0.35\textwidth]{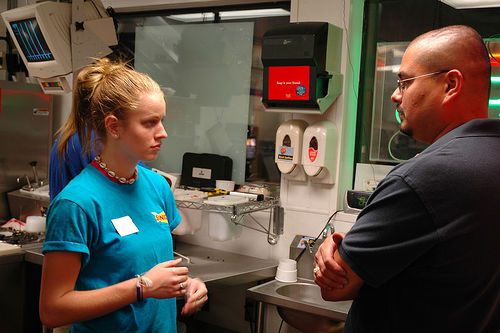}
\caption{Image 8063007 from the Flickr30K dataset.}
\label{fig:girlboss}
\end{figure}

What this assumption overlooks is the amount of \emph{interpretation} or \emph{recontextualization} carried out by the annotators. Let us take a concrete example. Figure \ref{fig:girlboss} shows an image from the Flickr30K dataset.

This image comes with the five descriptions below. All but the first one contain information that cannot come from the image alone. Relevant parts are highlighted in \textbf{bold}:

\begin{enumerate}
\item A blond girl and a bald man with his arms crossed are standing inside looking at each other.
\item A \textbf{worker} is \textbf{being scolded} by her \textbf{boss} in a \textbf{stern lecture}.
\item A \textbf{manager talks to an employee about job performance}.
\item A hot, blond girl \textbf{getting criticized by her boss}.
\item Sonic employees \textbf{talking about work}.
\end{enumerate}

We need to understand that the descriptions in the Flickr30K dataset are subjective descriptions of events. This can be a good thing: the descriptions tell us what are the salient parts of each image to the average human annotator. So the two humans in Figure \ref{fig:girlboss} are relevant, but the two soap dispensers are not. But subjectivity can also result in \emph{stereotypical} descriptions, in this case suggesting that the male is more likely to be the manager, and the female is more likely to be the subordinate. \newcite{rashtchian2010collecting} do note that some descriptions are speculative in nature, which they say hurts the accuracy and the consistency of the descriptions. But the problem is not with the lack of consistency here. Quite the contrary: the problem is that stereotypes may be pervasive enough for the data to be consistently biased. And so language models trained on this data may propagate harmful stereotypes, such as the idea that women are less suited for leadership positions.

This paper aims to give an overview of linguistic bias and unwarranted inferences resulting from stereotypes and prejudices. I will build on earlier work on linguistic bias in general \cite{beukeboom2014mechanisms}, providing examples from the Flickr30K data, and present a taxonomy of unwarranted inferences. Finally, I will discuss several methods to analyze the data in order to detect biases.\footnote{The Flickr30K data also contains examples where annotators judge the subjects of the images on their looks. E.g.\ description \#4 above calling the girl in the image \emph{hot}. Analyzing this judgmental language goes beyond the scope of this paper.}


\section{Stereotype-driven descriptions}
Stereotypes are ideas about how other (groups of) people commonly behave and what they are likely to do. These ideas guide the way we talk about the world. I distinguish two kinds of verbal behavior that result from stereotypes: (i) linguistic bias, and (ii) unwarranted inferences. The former is discussed in more detail by \newcite{beukeboom2014mechanisms}, who defines linguistic bias as ``a systematic asymmetry in word choice as a function of the social category to which the target belongs.'' So this bias becomes visible through the \emph{distribution} of terms used to describe entities in a particular category. Unwarranted inferences are the result of speculation about the image; here, the annotator goes beyond what can be glanced from the image and makes use of their knowledge and expectations about the world to provide an overly specific description. Such descriptions are directly identifiable as such, and in fact we have already seen four of them (descriptions 2--5) discussed earlier. 

\subsection{Linguistic bias}
Generally speaking, people tend to use more concrete or specific language when they have to describe a person that does not meet their expectations. \newcite{beukeboom2014mechanisms} lists several linguistic `tools' that people use to mark individuals who deviate from the norm. I will mention two of them.\footnote{Examples given are also due to \cite{beukeboom2014mechanisms}.}

\begin{description}[leftmargin=0cm]
\item[Adjectives] One well-studied example \cite{stahlberg2007representation,romaine2001corpus} is sexist language, where the sex of a person tends to be mentioned more frequently if their role or occupation is inconsistent with `traditional' gender roles (e.g.\ \emph{female surgeon, male nurse}). Beukeboom also notes that adjectives are used to create ``more narrow labels [or subtypes] for individuals who do not fit with general social category expectations'' (p. 3). E.g.\ \emph{tough woman} makes an exception to the `rule' that women aren't considered to be tough.

\item[Negation] can be used when prior beliefs about a particular social category are violated, e.g.\ \emph{The garbage man was not stupid.} See also \cite{beukeboom2010negation}.
\end{description}

These examples are similar in that the speaker has to put in additional effort to mark the subject for being unusual. But they differ in what \emph{we} can conclude about the speaker, especially in the context of the Flickr30K data. Negations are much more overtly displaying the annotator's prior beliefs. When one annotator writes that \emph{A little boy is eating pie \textbf{without} utensils} (image 2659046789), this immediately reveals the annotator's normative beliefs about the world: pie should be eaten \emph{with} utensils. But when another annotator talks about \emph{a girls basketball game} (image 8245366095), this cannot be taken as an indication that the annotator is biased about the gender of basketball players; they might just be helpful by providing a detailed description. In section 3 I will discuss how to establish whether or not there is any bias in the data regarding the use of adjectives.

\subsection{Unwarranted inferences}
Unwarranted inferences are statements about the subject(s) of an image that go beyond what the visual data alone can tell us. They are based on additional assumptions about the world. After inspecting a subset of the Flickr30K data, I have grouped these inferences into six categories (image examples between parentheses):

\begin{description}[leftmargin=0cm]
\item[Activity] We've seen an example of this in the introduction, where the `manager' was said to be \textit{talking about job performance} and \textit{scolding [a worker] in a stern lecture} (8063007).

\item[Ethnicity] Many dark-skinned individuals are called \emph{African-American} regardless of whether the picture has been taken in the USA or not (4280272). And people who look Asian are called Chinese (1434151732) or Japanese (4834664666).

\item[Event] In image 4183120 (Figure \ref{fig:watchingthegame}), people sitting at a gym are said to be watching a game, even though there could be any sort of event going on. But since the location is so strongly associated with sports, crowdworkers readily make the assumption.

\begin{figure}
\centering
\includegraphics[width=0.35\textwidth]{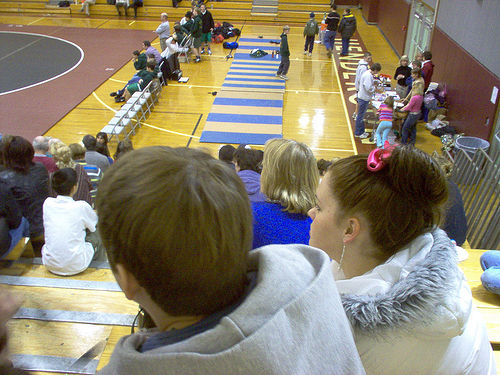}
\caption{Image 4183120 from the Flickr30K dataset.}
\label{fig:watchingthegame}
\end{figure}

\item[Goal] Quite a few annotations focus on explaining the \emph{why} of the situation. For example, in image 3963038375 a man is fastening his climbing harness \emph{in order to have some fun}. And in an extreme case, one annotator writes about a picture of a dancing woman that \textit{the school is having a special event in order to show the american culture on how other cultures are dealt with in parties} (3636329461). This is reminiscent of the Stereotypic Explanatory Bias \cite[SEB]{sekaquaptewa2003stereotypic}, which refers to ``the tendency to provide relatively more explanations in descriptions of stereotype inconsistent, compared to consistent behavior'' \cite[p. 5]{beukeboom2010negation}. So in theory, odd or surprising situations should receive more explanations, since a description alone may not make enough sense in those cases, but it is beyond the scope of this paper to test whether or not the Flickr30K data suffers from the SEB.

\item[Relation] Older people with children around them are commonly seen as parents (5287405), small children as siblings (205842), men and women as lovers (4429660), groups of young people as friends (36979).

\item[Status/occupation] Annotators will often guess the status or occupation of people in an image. Sometimes these guesses are relatively general (e.g. college-aged people being called \emph{students} in image 36979), but other times these are very specific (e.g.\ a man in a workshop being called a \emph{graphics designer}, 5867606). 
\end{description} 

\section{Detecting stereotype-driven descriptions}
In order to get an idea of the kinds of stereotype-driven descriptions that are in the Flickr30K dataset, I made a browser-based annotation tool that shows both the images and their associated descriptions.\footnote{\label{imageviewer}Code and data is available on GitHub: \url{https://github.com/evanmiltenburg/Flickr30K-Image-Viewer}} You can simply leaf through the images by clicking `Next' or `Random' until you find an interesting pattern. 

\subsection{Ethnicity/race}
One interesting pattern is that the ethnicity/race of babies doesn't seem to be mentioned \emph{unless} the baby is black or asian. In other words: white seems to be the default, and others seem to be marked. How can we tell whether or not the data is actually biased? 

We don't know whether or not an entity belongs to a particular social class (in this case: ethnic group) until it is marked as such. But we can approximate the proportion by looking at all the images where the annotators have used a marker (in this case: adjectives like \emph{black, white, asian}), and for those images count how many descriptions (out of five) contain a marker. This gives us an \emph{upper bound} that tells us how often ethnicity is indicated by the annotators. Note that this upper bound lies somewhere between 20\% (one description) and 100\% (5 descriptions). Figure \ref{table:babies} presents count data for the ethnic marking of babies. It includes two false positives (talking about a \emph{white baby stroller} rather than a \emph{white baby}). In the Asian group there is an additional complication: sometimes the mother gets marked rather than the baby. E.g.\ \textit{An Asian woman holds a baby girl.} I have counted these occurrences as well.

\begin{table}
{\small
\begin{tabular}{llr}
\toprule
\textbf{Asian} & \multicolumn{2}{r}{Average 60\%}\\
\midrule
2339632913 & Asian child/baby & 2\\
3208987435 & Asian baby, Asian/oriental woman & 3\\
7327356514 & Asian girl/baby, Asian/oriental woman & 4\\
\midrule
\textbf{Black} & \multicolumn{2}{r}{Average 40\%}\\
\midrule
1319788022 & African-American (AA)/black baby & 3\\
149057633 & African/AA child, black baby & 3\\
3217909454 & Dark-skinned baby & 1\\
3614582606 &	AA baby & 1\\
\midrule
\textbf{White} &  \multicolumn{2}{r}{Average 20\%}\\		
\midrule
11034843 & White baby boy & 1\\
176230509 & White baby boy & 1\\
2058947638 & White baby & 1\\
3991342877 & White baby & 1\\
4592281294 & White baby stroller & FP\\
661546153 & White baby stroller & FP\\
442983801 & Fair-skinned baby & 1\\
\bottomrule
\end{tabular}}
\caption{Number of times ethnicity/race was mentioned per category, per image. The average is expressed as a percentage of the number of descriptions. Counts in the last column correspond to the number of descriptions containing an ethnic/racial marker. Images were found by looking for descriptions matching \texttt{\small(asian|white|black|African-American|skinned) baby}. I found two false positives, indicated with FP.}
\label{table:babies}
\end{table}

The numbers in Table \ref{table:babies} are striking: there seems to be a real, systematic difference in ethnicity marking between the groups. We can take one step further and look at all the 697 pictures with the word `baby' in it. If there turn out to be disproportionately many white babies, this strengthens the conclusion that the dataset is biased.\footnote{Of course this extra step does constitute an additional annotation effort, and it is fairly difficult to automate; one would have to train a classifier for each group that needs to be checked.}

I have manually categorized each of the baby images. There are 504 white, 66 asian, and 36 black babies. 73 images do not contain a baby, and 18 images do not fall into any of the other categories. While this does bring down the average number of times each category was marked, it also increases the contrast between white babies (who get marked in less than 1\% of the images) and asian/black babies (who get marked much more often). A next step would be to see whether these observations also hold for other age groups, i.e.\ children and adults.$^{\ref{imageviewer}}$


\subsection{Other methods}
It may be difficult to spot patterns by just looking at a collection of images. Another method is to tag all descriptions with part-of-speech information, so that it becomes possible to see e.g.\ which adjectives are most commonly used for particular nouns. One method readers may find particularly useful is to leverage the structure of Flickr30K Entities \cite{plummer2015flickr30k}. This dataset enriches Flickr30K by adding coreference annotations, i.e.\ which phrase in each description refers to the same entity in the corresponding image. I have used this data to create a coreference graph by linking all phrases that refer to the same entity. Following this, I applied Louvain clustering \cite{blondel2008fast} to the coreference graph, resulting in clusters of expressions that refer to similar entities. Looking at those clusters helps to get a sense of the enormous variation in referring expressions. To get an idea of the richness of this data, here is a small sample of the phrases used to describe beards (cluster 268): \textit{a scruffy beard; a thick beard; large white beard; a bubble beard; red facial hair; a braided beard; a flaming red beard.} In this case, `red facial hair' really stands out as a description; why not choose the simpler `beard' instead?\footnote{Code and data is available on GitHub: \url{https://github.com/evanmiltenburg/Flickr30k-clusters}}

\section{Discussion}
In the previous section, I have outlined several methods to manually detect stereotypes, biases, and odd phrases. 
Because there are many ways in which a phrase can be biased, it is difficult to automatically detect bias from the data. So how should we deal with stereotype-driven descriptions?

\begin{description}[leftmargin=0cm]

\item[Neutralizing stereotypes for production]
One way to move forward might be to work with multilingual data. \newcite{elliott2015multilingual} propose a model that generates image descriptions given data from multiple languages, in their case German and English. Multilingual, or better: multicultural data might force models to put less emphasis on features that are only salient to annotators from one particular country.

\item[Stereotypes and interpretation]
While stereotypes might be a problem for production, further study of cultural stereotyping might be beneficial to systems that have to interpret human descriptions and determine likely referents of those descriptions. E.g.\ knowing that \emph{baseball player} probably refers to a \emph{male} baseball player is very useful.

\item[Levels of describing an image] There is a large body of work in art, information science, library science and related fields dedicated to the description and categorization of images \cite{shatford1986analyzing,jaimes1999conceptual}. A common thread is that we can divide image description into multiple levels or stages, starting from concrete physical attributes up to abstract contextual information. These levels build on each other; we first have to recognize separate entities before we can reason about their relation. But recent neural network models like \cite{vinyals2015show} do not match this procedure. Rather, they are trained to create a direct mapping between images and their descriptions. With this paper, I hope to have shown that the Flickr30K dataset is \emph{layered}, reflecting not only the physical contents of the images, but also whether the images match the everyday expectations of the crowd. An interesting challenge would be for image description models to learn separate representations for both layers: the perceptual and the contextual.

\item[Representativeness] My argument here is not that we should explicitly remove bias from crowdsourced descriptions of images. This may result in normalising the data into a form that is less representative of actual human descriptions. I do, however, contend that we should accept that crowdsourced descriptions of images \textit{are} biased. Acknowledging this fact is an important step towards designing models that can accommodate data based on a mixture of facts \textit{and stereotypes} about the world.
\end{description}

\section{Conclusion}
This paper provided a taxonomy of stereotype-driven descriptions in the Flickr30K dataset. I have divided these descriptions into two classes: linguistic bias and unwarranted inferences. The former corresponds to the annotators' choice of words when confronted with an image that may or may not match their stereotypical expectancies. The latter corresponds to the tendency of annotators to go beyond what the physical data can tell us, and expand their descriptions based on their past experiences and knowledge of the world. Acknowledging these phenomena is important, because on the one hand it helps us think about what is learnable from the data, and on the other hand it serves as a warning: if we train and evaluate language models on this data, we are effectively teaching them to be biased. 

I have also looked at methods to detect stereotype-driven descriptions, but due to the richness of language it is difficult to find an automated measure. Depending on whether your goal is production or interpretation, it may either be useful to suppress or to emphasize biases in human language. Finally, I have discussed stereotyping behavior as the addition of a contextual layer on top of a more basic description. This raises the question what kind of descriptions we would like our models to produce.

\section{Acknowledgments}
Thanks to Piek Vossen and Antske Fokkens for discussion, and to Desmond Elliott and an anonymous reviewer for comments on an earlier version of this paper. This research was supported by the Netherlands Organization for Scientific Research (NWO) via the Spinoza-prize awarded to Piek Vossen (SPI 30-673, 2014-2019).

\section{Bibliographical references}
\bibliographystyle{lrec2016}
\bibliography{image_bias_refs}
\end{document}